\documentclass[10pt,twocolumn,letterpaper]{article}
\pdfoutput=1

\makeatletter
\def\@maketitle
   {
   \newpage
   \null
   \vskip .375in
   \begin{center}
      {\Large \bf \@title \par}
      \vspace*{24pt}
      {
      \large
      \lineskip .5em
      \begin{tabular}[t]{c}
         \@author
      \end{tabular}
      \par
      }
      \vskip .5em
      \vspace*{12pt}
   \end{center}
   }
\makeatother

\def\abstract
   {%
   \centerline{\large\bf Abstract}%
   \vspace*{12pt}%
   \it%
   }

\def\affiliation#1{\gdef\@affiliation{#1}} \gdef\@affiliation{}

\makeatletter
\newlength{\@ctmp}
\newlength{\@figindent}
\long\def\@makecaption#1#2{
   \setbox\@tempboxa\hbox{\small \noindent #1.~#2}
   \setlength{\@ctmp}{\hsize}
   \addtolength{\@ctmp}{-\@figindent}\addtolength{\@ctmp}{-\@figindent}
   \ifdim \wd\@tempboxa >\@ctmp
      {\small #1.~#2\par}
   \else
      \hbox to\hsize{\hfil\box\@tempboxa\hfil}
  \fi}
  \makeatother

\font\elvbf  = ptmb scaled 1100

\makeatletter
\def\mysection{\@startsection {section}{1}{\z@}
   {10pt plus 2pt minus 2pt}{7pt} {\large\bf}}
\def\myssect#1{\mysection*{#1}}
\def\mysect#1{\mysection{\hskip -1em.~#1}}
\def\section{\@ifstar\myssect\mysect}

\def\mysubsection{\@startsection {subsection}{2}{\z@}
   {8pt plus 2pt minus 2pt}{6pt} {\elvbf}}
\def\myssubsect#1{\mysubsection*{#1}}
\def\mysubsect#1{\mysubsection{\hskip -1em.~#1}}
\def\subsection{\@ifstar\myssubsect\mysubsect}
\makeatother

\usepackage{times}
\usepackage{epsfig}
\usepackage{graphicx}
\usepackage{amsmath}
\usepackage{amssymb}

\usepackage{algorithm}
\usepackage{algpseudocode}
\usepackage{xfrac} 
\usepackage{tabularx} 
\usepackage{fancyhdr}

\fancypagestyle{plain}{

\fancyhf{}
\fancyhead[C]{MANUSCRIPT IN PRESS 2015 INTERNATIONAL CONFERENCE ON COMPUTER VISION}
}

\makeatletter\let\captiontemp\@makecaption\makeatother
\usepackage[font=footnotesize]{subcaption}
\makeatletter\let\@makecaption\captiontemp\makeatother

\usepackage[pagebackref=true,breaklinks=true,colorlinks,bookmarks=false]{hyperref}

\graphicspath{{./images/}}

\DeclareMathOperator*{\argmax}{arg\,max}
\DeclareMathOperator*{\argmin}{arg\,min}

\makeatletter
\let\OldStatex\Statex
\renewcommand{\Statex}[1][3]{
  \setlength\@tempdima{\algorithmicindent}
  \OldStatex\hskip\dimexpr#1\@tempdima\relax}
\makeatother

\newcommand{\etal}{\mbox{\emph{et al.\ }}}

\begin{document}

\setlength{\abovedisplayskip}{9.0pt plus 2.0pt minus 5.0pt}
\setlength{\belowdisplayskip}{9.0pt plus 2.0pt minus 5.0pt}

\title{An Adaptive Data Representation for Robust Point-Set Registration and Merging}

\author{Dylan Campbell and Lars Petersson\\
Australian National University\,\,\,\,\,\,\,\,\,\,National ICT Australia (NICTA)%
\thanks{\tiny NICTA is funded by the Australian Government through the Department of Communications and the Australian Research Council through the ICT Centre of Excellence Program.}\\
{\tt\small \{dylan.campbell,lars.petersson\}@nicta.com.au}
}

\maketitle

\begin{abstract}

This paper presents a framework for rigid point-set registration and merging using a robust continuous data representation. Our point-set representation is constructed by training a one-class support vector machine with a Gaussian radial basis function kernel and subsequently approximating the output function with a Gaussian mixture model. We leverage the representation's sparse parametrisation and robustness to noise, outliers and occlusions in an efficient registration algorithm that minimises the $L_2$ distance between our support vector--parametrised Gaussian mixtures. In contrast, existing techniques, such as Iterative Closest Point and Gaussian mixture approaches, manifest a narrower region of convergence and are less robust to occlusions and missing data, as demonstrated in the evaluation on a range of 2D and 3D datasets. Finally, we present a novel algorithm, GMMerge, that parsimoniously and equitably merges aligned mixture models, allowing the framework to be used for reconstruction and mapping.

\end{abstract}

\vspace{-6pt}
\section{Introduction}
\label{sec:introduction}

Point-set registration, the problem of finding the transformation that best aligns one point-set with another, is fundamental in computer vision, robotics, computer graphics and medical imaging.
A general-purpose point-set registration algorithm operates on unstructured point-sets and may not assume other information is available, such as labels or mesh structure.
Applications include merging multiple partial scans into a complete model~\cite{huber2003fully}; using registration results as fitness scores for object recognition~\cite{belongie2002shape}; registering a view into a global coordinate system for sensor localisation~\cite{nuchter20076d};
and finding relative poses between sensors~\cite{yang2013single}.

The dominant solution is the Iterative Closest Point (ICP) algorithm~\cite{besl1992method} and variants due to its conceptual simplicity, usability and good performance in practice. However, these are local techniques that are very susceptible to local minima and outliers and require a significant amount of overlap between point-sets.
To mitigate the problem of local minima, other solutions have widened the region of convergence~\cite{fitzgibbon2003robust}, performed heuristic global search~\cite{sandhu2010point}, used feature-based coarse alignment~\cite{rusu2009fast} or used branch-and-bound techniques to find the global minimum~\cite{yang2013goicp}.

Our method widens the region of convergence and is robust to occlusions and missing data, such as those arising when an object is viewed from different locations. The central idea is that the robustness of registration is dependent on the data representation used. We present a framework for robust point-set registration and merging using a continuous data representation, a Support Vector--parametrised Gaussian Mixture (SVGM). A discrete point-set is mapped to the continuous domain by training a Support Vector Machine (SVM) and mapping it to a Gaussian Mixture Model (GMM).
Since an SVM is parametrised by a sparse intelligently-selected subset of data points, an SVGM is compact and robust to noise, fragmentation and occlusions~\cite{nguyen2013support}, crucial qualities for efficient and robust registration.
The motivation for a continuous representation is that a typical scene comprises a single, seldom-disjoint continuous surface, which cannot be fully modelled by a discrete point-set sampled from the scene.

Our Support Vector Registration (SVR) algorithm minimises an objective function based on the $L_2$ distance between SVGMs.
Unlike the benchmark GMM registration algorithm GMMReg~\cite{jian2011robust}, SVR uses an adaptive and sparse representation with non-uniform and data-driven mixture weights, enabling faster performance and improving the robustness to outliers, occlusions and partial overlap.

\begin{figure*}[!t]
\centering
\setlength{\unitlength}{496.85625pt} 
\begin{picture}(1, 0.32)(0, 0) 
\put(0.00, 0){\frame{\includegraphics[width=0.16\textwidth, trim=0 0 3pt 3pt, clip]{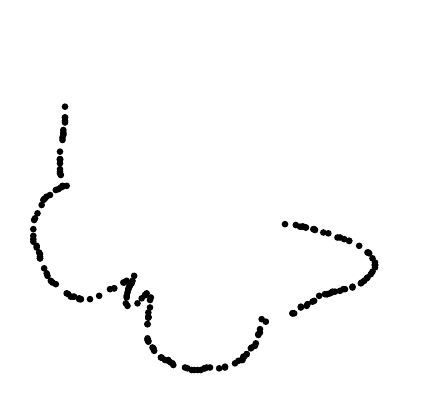}}}
\put(0.21, 0){\frame{\includegraphics[width=0.16\textwidth, trim=0 0 3pt 3pt, clip]{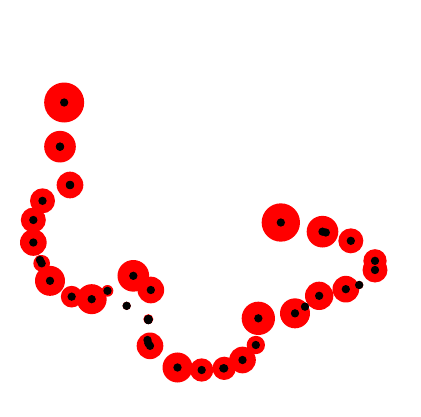}}}
\put(0.42, 0){\frame{\includegraphics[width=0.16\textwidth, trim=0 0 3pt 3pt, clip]{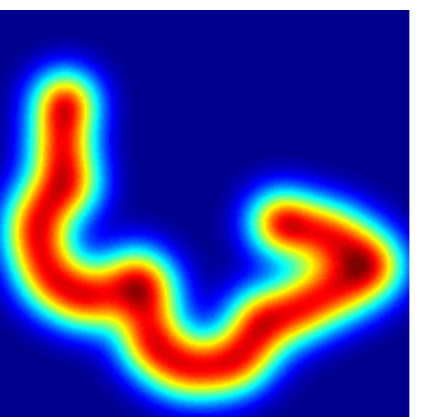}}}
\put(0.63, 0){\frame{\includegraphics[width=0.16\textwidth, trim=0 0 3pt 3pt, clip]{butterfly_gy.pdf}}}

\put(0.00, 0.16){\frame{\includegraphics[width=0.16\textwidth, trim=0 0 3pt 3pt, clip]{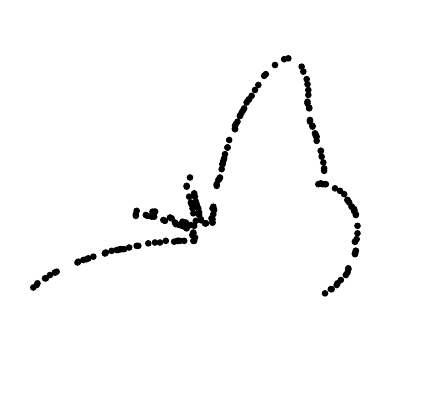}}}
\put(0.21, 0.16){\frame{\includegraphics[width=0.16\textwidth, trim=0 0 3pt 3pt, clip]{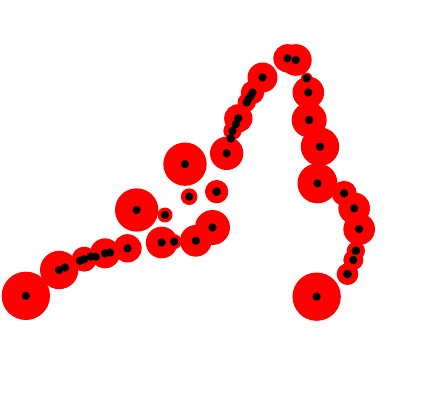}}}
\put(0.42, 0.16){\frame{\includegraphics[width=0.16\textwidth, trim=0 0 3pt 3pt, clip]{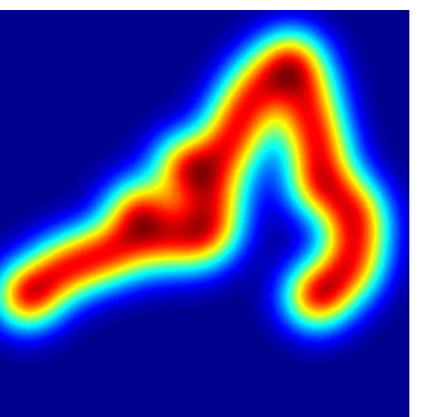}}}
\put(0.63, 0.16){\frame{\includegraphics[width=0.16\textwidth, trim=0 0 3pt 3pt, clip]{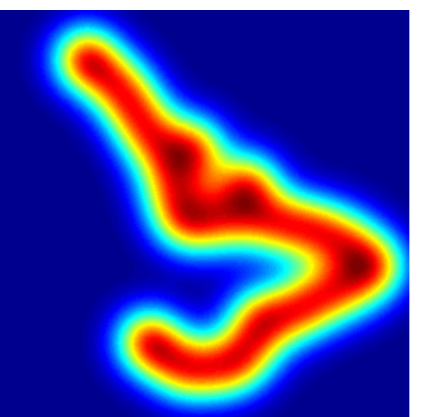}}}

\put(0.84, 0.08){\frame{\includegraphics[width=0.16\textwidth, trim=0 0 3pt 3pt, clip]{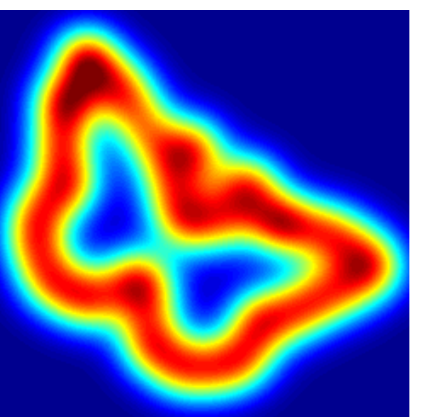}}}

\thicklines
\put(0.16, 0.08){\vector(1,0){0.05}}
\put(0.37, 0.08){\vector(1,0){0.05}}
\put(0.58, 0.08){\vector(1,0){0.05}}
\put(0.79, 0.08){\vector(1,1){0.05}}


\put(0.16, 0.24){\vector(1,0){0.05}}
\put(0.37, 0.24){\vector(1,0){0.05}}
\put(0.58, 0.24){\vector(1,0){0.05}}
\put(0.79, 0.24){\vector(1,-1){0.05}}

\put(0.00, 0.32){\makebox(0.16, 0.02)[t]{Point-Set}}
\put(0.21, 0.32){\makebox(0.16, 0.02)[t]{SVM}}
\put(0.42, 0.32){\makebox(0.16, 0.02)[t]{Misaligned SVGM}}
\put(0.63, 0.32){\makebox(0.16, 0.02)[t]{Aligned SVGM}}

\put(0.84, 0.24){\makebox(0.16, 0.02)[t]{Merged SVGM}}

\put(0.16, 0.24){\makebox(0.05, 0.02)[t]{(a)}}
\put(0.37, 0.24){\makebox(0.05, 0.02)[t]{(b)}}
\put(0.58, 0.24){\makebox(0.05, 0.02)[t]{(c)}}
\put(0.79, 0.23){\makebox(0.05, 0.02)[t]{(d)}}

\end{picture}
\caption{Robust point-set registration and merging framework. An $n$D point-set is represented as an SVGM by training a one-class SVM (a) and then mapping it to a GMM (b). The SVR algorithm is used to minimise the $L_2$ distance between two SVGMs in order to align the densities (c). Finally, the GMMerge algorithm is used to parsimoniously fuse the two mixtures. The SVMs are visualised as support vector points scaled by mixture weight and the SVGMs are coloured by probability value. Best viewed in colour.}
\label{fig:framework}
\end{figure*}

Finally, we propose a novel merging algorithm, GMMerge, that parsimoniously and equitably merges aligned mixtures. Merging SVGM representations is useful for applications where each point-set may contain unique information, such as reconstruction and mapping. Our registration and merging framework is visualised in Figure~\ref{fig:framework}.

\section{Related Work}
\label{sec:related_work}

The large volume of work published on ICP, its variants and other registration techniques precludes a comprehensive list, however the reader is directed to recent surveys on ICP variants~\cite{pomerleau2013comparing} and 3D point-set and mesh registration techniques~\cite{tam2013registration} for additional background. Of relevance to our work are extensions that improve its occlusion robustness, such as trimming~\cite{chetverikov2005robust}. Local methods that seek to improve upon ICP's basin of convergence and sensitivity to outliers include LM-ICP~\cite{fitzgibbon2003robust}, which uses a distance transform to optimise the ICP error without establishing explicit point correspondences.

Another family of approaches, to which ours belongs, is based on the Gaussian Mixture Model (GMM) and show an improved robustness to poor initialisations, noise and outliers. Notable GMM algorithms for rigid and non-rigid registration include Robust Point Matching~\cite{chui2003new}, using soft assignment and deterministic annealing, Coherent Point Drift~\cite{myronenko2010point}, Kernel Correlation~\cite{tsin2004correlation} and GMMReg~\cite{jian2011robust}. The latter two do not establish explicit point correspondences and both minimise a distance measure between mixtures. GMMReg~\cite{jian2011robust} defines an equally-weighted Gaussian at every point in the set with identical and isotropic covariances and minimises the $L_2$ distance between mixtures.
The Normal Distributions Transform (NDT) algorithm~\cite{magnusson2007scan} is a similar method, defining Gaussians for every cell in a grid discretisation and estimating full data-driven covariances, like~\cite{xiong2013study}. Unlike our method, however, it imposes external structure on the scene and uses uniform mixture weights.

In contrast, globally-optimal techniques avoid local minima by searching the entire transformation space. Existing 3D methods~\cite{li20073d,yang2013goicp} are often very slow or make restrictive assumptions about the point-sets or transformations.
There are also many heuristic or stochastic methods for global alignment that are not guaranteed to converge, such as particle filtering~\cite{sandhu2010point}, genetic algorithms~\cite{silva2005precision} and feature-based alignment~\cite{rusu2009fast}.
A recent example is \textsc{Super 4PCS}, a four-points congruent sets method that exploits a clever data structure to achieve linear-time performance~\cite{mellado2014super}.

The rest of the paper is organised as follows: we present the SVGM representation, its properties and implementation in Section~\ref{sec:point-set_representation}, we develop a robust framework for SVGM registration in Section~\ref{sec:svmreg}, we propose an algorithm for merging SVGMs in Section~\ref{sec:merging}, we experimentally demonstrate the framework's effectiveness in Section~\ref{sec:results} and we discuss the results and conclude in Sections~\ref{sec:discussion} and~\ref{sec:conclusion}.

 \section{Adaptive Point-Set Representation}
 \label{sec:point-set_representation}

A central idea of our work is that the robustness of point-set registration is dependent on the data representation used. Robustness to occlusions or missing data, more so than noise, is of primary concern, because point-sets rarely overlap completely, such as when an object is sampled from a different sensor location.
Another consideration is the class of optimisation problem a particular representation admits. Framing registration as a continuous optimisation problem involving continuous density functions may make it more tractable than the equivalent discrete problem~\cite{jian2011robust}.
Consequently, we represent discrete point-sets with Gaussian Mixture Models (GMMs). Crucially, we first train a Support Vector Machine (SVM) and then transform this into a GMM. Since the output function of the SVM only involves a sparse subset of the data points, the representation is compact and robust to noise, fragmentation and occlusions~\cite{nguyen2013support}, attributes that persist through the GMM transformation.

\subsection{One-Class Support Vector Machine}
\label{sec:one-class_svm}

The output function of an SVM can be used to approximate the surface described by noisy and incomplete point-set data, providing a continuous implicit surface representation.
Nguyen and Porikli \cite{nguyen2013support} demonstrated that this representation is robust to noise, fragmentation, missing data and other artefacts for 2D shapes, with the same behaviour expected in 3D.
An SVM classifies data by constructing a hyperplane that separates data of two different classes, maximising the margin between the classes while allowing for some mislabelling \cite{cortes1995support}. Since point-set data contains only positive examples, one-class SVM \cite{scholkopf2001estimating} can be used to find the hyperplane that maximally separates the data points from the origin or viewpoint in feature space. The training data is mapped to a higher-dimensional feature space, where it may be linearly separable from the origin, with a non-linear kernel function.

The output function $f(\mathbf{x})$ of one-class SVM is given by
\begin{equation}
\label{eqn:svm_output_function}
f(\mathbf{x}) = \sum_{i = 1}^{\ell} \alpha_{i} K(\mathbf{x}_{i}, \mathbf{x}) - \rho
\end{equation}
where $\mathbf{x}_i$ are the point vectors, $\alpha_i$ are the weights, $\mathbf{x}$ is the input vector, $\rho$ is the bias, $\ell$ is the number of training samples and $K$ is the kernel function
that evaluates the inner product of data vectors mapped to a feature space.
We use a Gaussian Radial Basis Function (RBF) kernel
\begin{equation}
\label{eqn:GRBF}
K(\mathbf{x}_{i}, \mathbf{x}) = \exp \left(-\gamma \left\| \mathbf{x}_{i} - \mathbf{x} \right\|_2^2 \right)
\end{equation}
where $\gamma$ is the Gaussian kernel width.

The optimisation formulation in~\cite{scholkopf2001estimating} has a parameter $\nu \in (0, 1]$ that controls the trade-off between training error and model complexity. It is a lower bound on the fraction of support vectors and an upper bound on the misclassification rate~\cite{scholkopf2001estimating}. The data points with non-zero weights $\alpha_{i}^{\mathrm{SV}}$ are the support vectors $\mathbf{x}_{i}^{\mathrm{SV}} \in \{\mathbf{x}_{i} : \alpha_{i} > 0, i = 1, \ldots, \ell\}$.

We estimate the kernel width $\gamma$ automatically for each point-set by noting that it is inversely proportional to the square of the scale $\sigma$. For an $\ell \times D$ point-set $\mathbf{X}$ with mean $\bar{\mathbf{x}}$, the estimated scale $\hat{\sigma}$ is proportional to the $2D$th root of the generalised variance
\begin{equation}
\label{eqn:sigma_hat}
\hat{\sigma} \propto
\left|
\frac{1}{\ell - 1} (\mathbf{X} - \mathbf{1} \bar{\mathbf{x}}^{\intercal})^{\intercal} (\mathbf{X} - \mathbf{1} \bar{\mathbf{x}}^{\intercal})
\right|
^{\sfrac{1}{2D}}.
\end{equation}
If a training set is available, better performance can be achieved by finding $\gamma$ using cross-validation, imposing a constraint on the registration accuracy and searching in the neighbourhood of $\sfrac{1}{2\hat{\sigma}^2}$.

\subsection{Gaussian Mixture Model Transformation}
\label{sec:gmm}

In order to make use of the trained SVM for point-set registration, it must first be approximated as a GMM. We use the transformation identified by Deselaers \etal \cite{deselaers2010object} to represent the SVM in the framework of a GMM, without altering the decision boundary.
A GMM converted from an SVM will necessarily optimise classification performance instead of data representation, since SVMs are discriminative models, unlike standard generative GMMs.
This allows it to discard redundant data and reduces its susceptibility to varying point densities, which are prevalent in real datasets.

The decision function of an SVM with a Gaussian RBF kernel can be written as
\begin{equation}
\label{eqn:svm_decision}
r(\mathbf{x}) = \argmax_{k \in \{ -1,1\}} \left\{ \sum_{i = 1}^{\ell^{\mathrm{SV}}} k\alpha_{i}^{\mathrm{SV}} \mathrm{e}^{-\gamma \left\| \mathbf{x}_{i}^{\mathrm{SV}} - \mathbf{x} \right\|_2^2} - k\rho \right\}
\end{equation}
where $\ell^{\mathrm{SV}}$ is the number of support vectors and $k$ is the class, positive for inliers and negative otherwise for one-class SVM. The GMM decision function can be written as
\begin{equation}
\label{eqn:gmm_decision}
r'(\mathbf{x}) = \argmax_{k \in \{ -1,1\}} \left\{ \sum_{i = 1}^{I_{k}} p(k)p(i|k) \mathcal{N} \left( \mathbf{x} \middle| \boldsymbol{\mu}_{ki} , \sigma_{k}^{2} \right) \right\}
\end{equation}
where $I_{k}$ is the number of clusters for class $k$, $p(k)$ is the prior probability of class $k$, $p(i|k)$ is the cluster weight of the $i$th cluster of class $k$ and $\mathcal{N} \left( \mathbf{x} \middle| \boldsymbol{\mu}_{ki} , \sigma_{k}^{2} \right)$ is the Gaussian representing the $i$th cluster of class $k$ with mean $\boldsymbol{\mu}_{ki}$ and variance $\sigma_{k}^2$, given by
\begin{equation}
\label{eqn:gmm_normal_dist}
\mathcal{N} \left( \mathbf{x} \middle| \boldsymbol{\mu}_{ki} , \sigma_{k}^{2} \right) = \frac{1}{(2 \pi \sigma_{k}^{2})^{\sfrac{D}{2}}} \exp \left(- \frac{\left\| \mathbf{x} - \boldsymbol{\mu}_{ki} \right\|_{2}^{2}}{2\sigma_{k}^{2}} \right).
\end{equation}

Noting the similarity of (\ref{eqn:svm_decision}) and (\ref{eqn:gmm_decision}), the mapping
\begin{align}
\boldsymbol{\mu}_{ki} &= 
\begin{cases}
\mathbf{x}_{i}^{\mathrm{SV}} & \text{if } k = +1\\
\mathbf{0} & \text{else}
\end{cases}
\label{eqn:transform_mu} \\
\sigma_{k}^{2} &= 
\begin{cases}
\sfrac{1}{2\gamma} & \text{if } k = +1\\
N_{\infty} & \text{else}
\end{cases}
\label{eqn:transform_sigma} \\
\phi_{i} = p(k)p(i|k) &= 
\begin{cases}
\alpha_{i}^{\mathrm{SV}} (2\pi\sigma_{k}^{2})^{\sfrac{D}{2}} & \text{if } k = +1\\
\rho (2\pi\sigma_{k}^{2})^{\sfrac{D}{2}} & \text{else}
\end{cases}
\label{eqn:transform_p}
\end{align}
can be applied, 
where $\phi_{i}$ is the mixture weight, that is, the prior probability of the $i$th component. The bias term $\rho$ is approximated by an additional density given to the negative class with arbitrary mean, very high variance $N_{\infty}$ and a cluster weight proportional to $\rho$. We omit this term from the registration framework because it does not affect the optimisation. The resulting GMM is parametrised by
\begin{equation}
\label{eqn:gmm_set}
\mathcal{G} = \left\{ \boldsymbol{\mu}_{i} , \; \sigma^{2} , \; \phi_{i} \right\}_{i = 1}^{\ell^{\mathrm{SV}}}.
\end{equation}

\begin{figure}[!t]
\centering
\begin{subfigure}[]{0.32\columnwidth}
\includegraphics[width=\columnwidth]{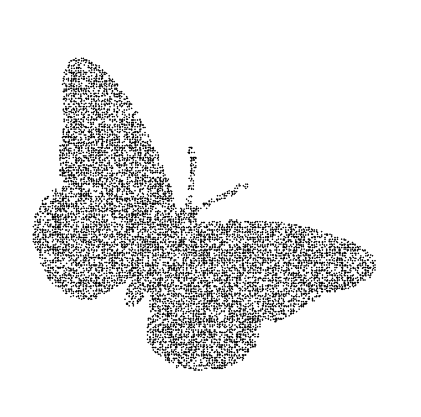}
\caption{Point-Set A}
\label{fig:butterfly_occlusion_A}
\end{subfigure}
\hfill
\begin{subfigure}[]{0.32\columnwidth}
\includegraphics[width=\columnwidth]{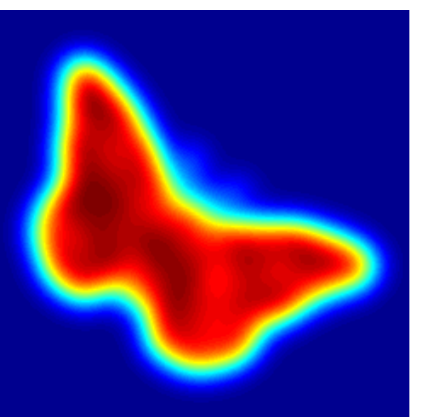}
\caption{KDE-GMM A}
\label{fig:butterfly_occlusion_kdeA}
\end{subfigure}
\hfill
\begin{subfigure}[]{0.32\columnwidth}
\includegraphics[width=\columnwidth]{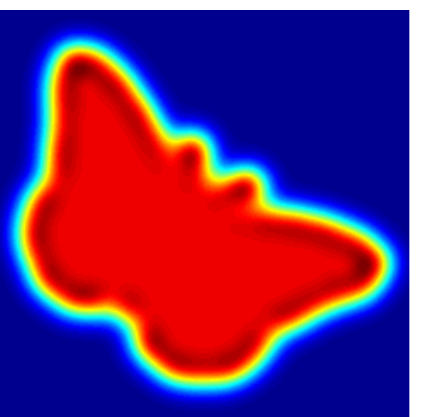}
\caption{SVGM A}
\label{fig:butterfly_occlusion_svmA}
\end{subfigure}

\begin{subfigure}[]{0.32\columnwidth}
\includegraphics[width=\columnwidth]{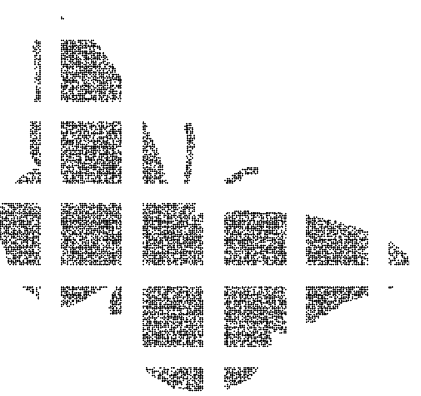}
\caption{Point-Set B}
\label{fig:butterfly_occlusion_B}
\end{subfigure}
\hfill
\begin{subfigure}[]{0.32\columnwidth}
\includegraphics[width=\columnwidth]{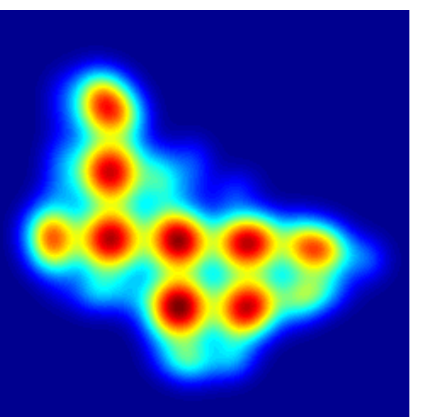}
\caption{KDE-GMM B}
\label{fig:butterfly_occlusion_kdeB}
\end{subfigure}
\hfill
\begin{subfigure}[]{0.32\columnwidth}
\includegraphics[width=\columnwidth]{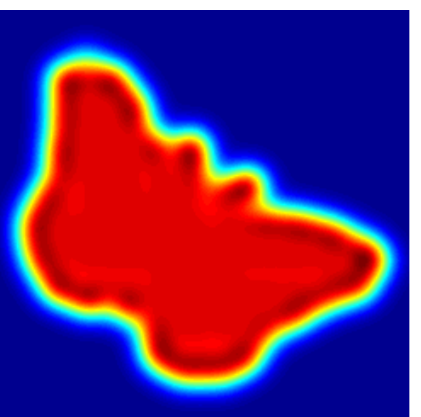}
\caption{SVGM B}
\label{fig:butterfly_occlusion_svmB}
\end{subfigure}
\caption{The effect of significant occlusion on two point-set representations, using the same parameters for both. Our SVGM representation is, qualitatively, almost identical when occluded (\subref{fig:butterfly_occlusion_svmB}) and unoccluded (\subref{fig:butterfly_occlusion_svmA}), whereas the fixed-bandwidth KDE representation is much less robust to occlusion (\subref{fig:butterfly_occlusion_kdeB}).
Best viewed in colour.}
\label{fig:butterfly_occlusion}
\end{figure}

While we transform an SVM into a GMM, there are many other ways to construct a GMM from point-set data. Kernel Density Estimation (KDE) with identically-weighted Gaussian densities has frequently been used for this purpose, including fixed-bandwidth KDE with isotropic covariances~\cite{jian2011robust,detry2009probabilistic}, variable-bandwidth KDE with non-identical covariances~\cite{comaniciu2003algorithm} and non-isotropic covariance KDE ~\cite{xiong2013study}.
%
The primary disadvantage of these methods is that the number of Gaussian components is equal to the point-set size, which can be very large for real-world datasets. In contrast, our work intelligently selects a sparse subset of the data points to locate the Gaussian densities and weights them non-identically, making it more robust to occlusions and missing data, as demonstrated in Figure~\ref{fig:butterfly_occlusion}.

Expectation Maximisation (EM)~\cite{dempster1977maximum} can also be used to construct a GMM with fewer components than KDE. EM finds the maximum likelihood estimates of the GMM parameters, where the number of densities is specified a priori, unlike our method. To initialise the algorithm, the means can be chosen at random or using the k-means algorithm; or, an initial Gaussian can be iteratively split and re-estimated until the number of densities is reached~\cite{deselaers2010object}. However, deliberately inflating the number of components can be slow and sensitive to initialisation~\cite[p. 326]{scott2001kernels}.

\section{Support Vector Registration}
\label{sec:svmreg}

Once the point-sets are in mixture model form, the registration problem can be posed as minimising the distance between mixtures.
Like Jian and Vemuri \cite{jian2011robust}, we use the $L_2$ distance, which can be expressed in closed-form.
The $L_2 E$ estimator minimises the $L_2$ distance between densities and is known, counter-intuitively, to be inherently robust to outliers \cite{scott2001parametric}, unlike the maximum likelihood estimator that minimises the Kullback-Leibler divergence.

Let $\mathcal{X}$ be the moving model point-set, $\mathcal{Y}$ be the fixed scene point-set, $\mathcal{G}_{\mathcal{X}}$ and $\mathcal{G}_{\mathcal{Y}}$ be GMMs converted from SVMs trained on $\mathcal{X}$ and $\mathcal{Y}$ respectively, and $T(\mathcal{G},\boldsymbol{\theta})$ be the transformation model parametrised by $\boldsymbol{\theta}$. The $L_2$ distance between transformed $\mathcal{G}_{\mathcal{X}}$ and $\mathcal{G}_{\mathcal{Y}}$ is given by
\begin{equation}
\label{eqn:l2_distance}
D_{L_{2}}(\mathcal{G}_{\mathcal{X}}, \mathcal{G}_{\mathcal{Y}}, \boldsymbol{\theta}) = \int_{\mathbb{R}^{D}} \left( p\left( \mathbf{x} \middle| T(\mathcal{G}_{\mathcal{X}}, \boldsymbol{\theta}) \right) - p\left( \mathbf{x} \middle| \mathcal{G}_{\mathcal{Y}} \right) \right)^{2}\,\mathrm{d}\mathbf{x}
\end{equation}
where $p\left( \mathbf{x} \middle| \mathcal{G} \right)$ is the probability of observing a point $\mathbf{x}$ given a mixture model $\mathcal{G}$ with $\ell$ components, that is
\begin{equation}
\label{eqn:gmm_probability}
p\left( \mathbf{x} \middle| \mathcal{G} \right) = \sum_{i = 1}^{\ell} \phi_{i} \mathcal{N} \left( \mathbf{x} \middle| \boldsymbol{\mu}_{i} , \sigma^{2} \right).
\end{equation}
Expanding (\ref{eqn:l2_distance}), the last term is independent of $\boldsymbol{\theta}$ and the first term is invariant under rigid transformations. Both are therefore removed from the objective function. The middle term is the inner product of two Gaussian mixtures and has a closed form that can be derived by applying the identity
\begin{multline}
\label{eqn:gmm_identity}
\int_{\mathbb{R}^{D}} \mathcal{N} \left( \mathbf{x} \middle| \boldsymbol{\mu}_{1} , \sigma_{1}^{2} \right) \mathcal{N} \left( \mathbf{x} \middle| \boldsymbol{\mu}_{2} , \sigma_{2}^{2} \right) \,\mathrm{d}\mathbf{x}\\
= \mathcal{N} \left( \mathbf{0} \middle| \boldsymbol{\mu}_{1} - \boldsymbol{\mu}_{2} , \sigma_{1}^{2} + \sigma_{2}^{2} \right).
\end{multline}

Therefore, noting that $\sigma_{\mathcal{X}}^{2} = \sigma_{\mathcal{Y}}^{2}$ in our formulation, the objective function for rigid registration is defined as
\begin{equation}
\label{eqn:objective_function}
\begin{aligned}
f\left(\boldsymbol{\theta} \right) &= - \sum_{i = 1}^{m} \sum_{j = 1}^{n} \phi_{i, \mathcal{X}} \phi_{j, \mathcal{Y}} \mathcal{N} \left( \mathbf{0} \middle| \boldsymbol{\mu}_{i, \mathcal{X}}' - \boldsymbol{\mu}_{j, \mathcal{Y}} , 2\sigma^{2} \right)\\
\end{aligned}
\end{equation}
where $m$ and $n$ are the number of components in $\mathcal{G}_{\mathcal{X}}$ and $\mathcal{G}_{\mathcal{Y}}$ respectively and $\boldsymbol{\mu}_{i, \mathcal{X}}' = T(\boldsymbol{\mu}_{i, \mathcal{X}}, \boldsymbol{\theta})$. This can be expressed in the form of a discrete Gauss transform, which has a computational complexity of $\mathcal{O}(mn)$, or the fast Gauss transform \cite{greengard1991fast}, which scales as $\mathcal{O}(m + n)$.

The gradient vector is derived as in~\cite{jian2011robust}. Let $\mathbf{M}_0 = \left[ \boldsymbol{\mu}_{1,\mathcal{X}}, \dots , \boldsymbol{\mu}_{m,\mathcal{X}}\right]^{\intercal}$ be the $m \times D$ matrix of the means from $\mathcal{G}_{\mathcal{X}}$ and $\mathbf{M} = T(\mathbf{M}_0, \boldsymbol{\theta})$ be the transformed matrix, parametrised by $\boldsymbol{\theta}$. Using the chain rule, the gradient is $\frac{\partial f}{\partial \boldsymbol{\theta}} = \frac{\partial f}{\partial \mathbf{M}} \frac{\partial \mathbf{M}}{\partial \boldsymbol{\theta}}$. 
Let $\mathbf{G} = \frac{\partial f}{\partial \mathbf{M}}$ be an $m \times D$ matrix, which can be found while evaluating the objective function by
\begin{equation}
\label{eqn:gradient_G}
\mathbf{G}_{i} = - \frac{1}{2\sigma^2} \sum_{j = 1}^{m} f_{ij} \left( \boldsymbol{\mu}_{i, \mathcal{X}}' - \boldsymbol{\mu}_{j, \mathcal{Y}} \right)
\end{equation}
where $\mathbf{G}_{i}$ is the $i$th row of $\mathbf{G}$ and $f_{ij}$ is a summand of $f$.
For rigid motion, $\mathbf{M} = \mathbf{M}_0 \mathbf{R}^{\intercal} + \mathbf{t}$ where $\mathbf{R}$ is the rotation matrix and $\mathbf{t}$ is the translation vector. The gradients with respect to each motion parameter are given by
\begin{align}
\frac{\partial f}{\partial \mathbf{t}} &= \mathbf{G}^{\intercal} \mathbf{1}_{m}
\label{eqn:gradient_translation} \\
\frac{\partial f}{\partial r_{i}} &= \mathbf{1}_{D}^{\intercal} \left( \left( \mathbf{G}^{\intercal} \mathbf{M}_{0} \right) \circ \frac{\partial \mathbf{R}}{\partial r_{i}} \right) \mathbf{1}_{D}
\label{eqn:gradient_rotation}
\end{align}
where $\mathbf{1}_{i}$ is the $i$-dimensional column vector of ones, $\circ$ is the Hadamard
product and $r_{i}$ are the elements parametrising $\mathbf{R}$: rotation angle $\alpha$ for 2D and a unit quaternion for 3D.
For the latter, the quaternion is projected back to the space of valid rotations after each update by normalisation.

Since the objective function is smooth, differentiable and convex in the neighbourhood of the optimal motion parameters, gradient-based numerical optimisation methods can be used, such as nonlinear conjugate gradient or quasi-Newton methods. We use an interior-reflective Newton method \cite{coleman1996interior} since it is time and memory efficient and scales well.
However, since the objective function is non-convex over the search space, this approach is susceptible to local minima, particularly for large motions and point-sets with symmetries.
A multi-resolution approach can be adopted, increasing $\gamma$ at each iteration and initialising with the currently optimal transformation.
SVR is outlined in Algorithm~\ref{alg:SVR}.
\begin{algorithm}
\begin{algorithmic}[1]
\Require model point-set $\mathcal{X} = \{\mathbf{x}_{i} \}_{i = 1}^{\ell_{\mathcal{X}}}$, scene point-set $\mathcal{Y} = \{\mathbf{y}_{i} \}_{i = 1}^{\ell_{\mathcal{Y}}}$, transformation model $T$ parametrised by $\boldsymbol{\theta}$, initial parameter $\boldsymbol{\theta}_{0}$ such as the identity transformation

\Ensure locally optimal transformation parameter $\boldsymbol{\theta}^*$ such that $T(\mathcal{X}, \boldsymbol{\theta}^*)$ is best aligned with $\mathcal{Y}$

\State Select $\nu$ and $\gamma$ by estimation or cross-validation

\State Initialise transformation parameter: $\boldsymbol{\theta} \gets \boldsymbol{\theta}_{0}$

\Repeat

\State Train SVMs:
\Statex[1] $\mathcal{S}_{\mathcal{X}} = \left\{ \mathbf{x}_{i}^{\mathrm{SV}}, \; \alpha_{i,\mathcal{X}}^{\mathrm{SV}} \right\}_{i = 1}^{m} \gets \mathrm{trainSVM}(\mathcal{X}, \nu, \gamma)$
\Statex[1] $\mathcal{S}_{\mathcal{Y}} = \left\{ \mathbf{y}_{i}^{\mathrm{SV}}, \; \alpha_{i,\mathcal{Y}}^{\mathrm{SV}} \right\}_{i = 1}^{n} \gets \mathrm{trainSVM}(\mathcal{Y}, \nu, \gamma)$

\State Convert SVMs to GMMs using (\ref{eqn:transform_mu}), (\ref{eqn:transform_sigma}) and (\ref{eqn:transform_p}):
\Statex[1] $\mathcal{G}_{\mathcal{X}} = \left\{ \boldsymbol{\mu}_{i, \mathcal{X}} , \; \sigma^{2} , \; \phi_{i, \mathcal{X}} \right\}_{i = 1}^{m} \gets \mathrm{toGMM}(\mathcal{S}_{\mathcal{X}}, \gamma)$
\Statex[1] $\mathcal{G}_{\mathcal{Y}} = \left\{ \boldsymbol{\mu}_{i, \mathcal{Y}} , \; \sigma^{2} , \; \phi_{i, \mathcal{Y}} \right\}_{i = 1}^{n} \gets \mathrm{toGMM}(\mathcal{S}_{\mathcal{Y}}, \gamma)$

\State Optimise the objective function $f$~(\ref{eqn:objective_function}) using the \Statex[1] gradient~(\ref{eqn:gradient_translation}), (\ref{eqn:gradient_rotation}) with a trust region algorithm

\State Update the parameter $\boldsymbol{\theta} \gets \argmin_{\boldsymbol{\theta}} f\left(\boldsymbol{\theta} \right)$


\State Anneal: $\gamma \gets \delta\gamma$

\Until{change in $f$ or iteration number meets a condition}
\end{algorithmic}
\caption{Support Vector Registration (SVR): A robust algorithm for point-set registration using one-class SVM}
\label{alg:SVR}
\end{algorithm}


\section{Merging Gaussian Mixtures}
\label{sec:merging}

For an SVGM to be useful for applications where each point-set may contain unique information, such as mapping, an efficient method of merging two aligned mixtures is desirable.
A na\"{i}ve approach is to use a weighted sum of the Gaussian mixtures \cite{deselaers2010object}, however, this would result in an unnecessarily high number of components with substantial redundancy. Importantly, the probability of regions not observed in both point-sets would decrease, meaning that regions that are often occluded would disappear from the model as more mixtures were merged. While the time-consuming process of sampling the combined mixture and re-estimating it with EM would eliminate redundancy, it would not alleviate the missing data problem. The same applies to faster sample-free variational-Bayes approaches \cite{bruneau2010parsimonious}. Sampling (or merging the point-sets) and re-estimating an SVGM would circumvent this problem, since the discriminative framework of the SVM is insensitive to higher-density overlapping regions, but this is not time efficient.

\begin{figure*}[!t]
\centering
\begin{subfigure}[]{0.19\textwidth}
\includegraphics[width=\columnwidth]{butterfly_gx.pdf}
\caption{Aligned mixture $\mathcal{G}_{\mathcal{X}}$}
\label{fig:merge_gx}
\end{subfigure}
\hfill
\begin{subfigure}[]{0.19\textwidth}
\includegraphics[width=\columnwidth]{butterfly_gy.pdf}
\caption{Aligned mixture $\mathcal{G}_{\mathcal{Y}}$}
\label{fig:merge_gy}
\end{subfigure}
\hfill
\begin{subfigure}[]{0.19\textwidth}
\includegraphics[width=\columnwidth]{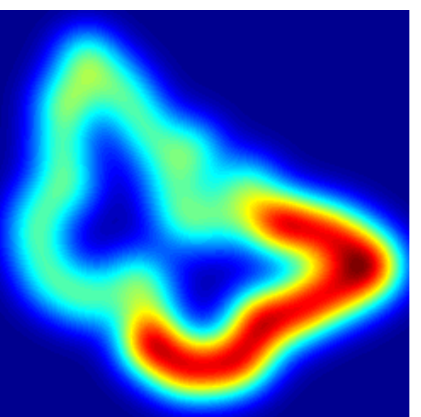}
\caption{Na\"{i}ve merge}
\label{fig:merge_naivemerge}
\end{subfigure}
\hfill
\begin{subfigure}[]{0.19\textwidth}
\includegraphics[width=\columnwidth]{butterfly_gmmerge.pdf}
\caption{GMMerge}
\label{fig:merge_gmmerge}
\end{subfigure}
\hfill
\begin{subfigure}[]{0.19\textwidth}
\includegraphics[width=\columnwidth]{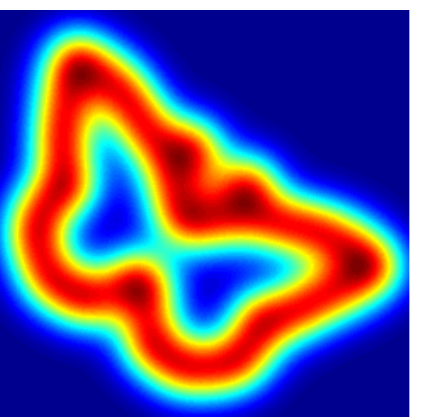}
\caption{Ground truth merge}
\label{fig:merge_gtmerge}
\end{subfigure}
\caption{Merging Gaussian mixtures (\subref{fig:merge_gx}) and (\subref{fig:merge_gy}) with a na\"{i}ve weighted sum (\subref{fig:merge_naivemerge}) and GMMerge (\subref{fig:merge_gmmerge}). The mixture produced by GMMerge is almost identical to the ground truth (\subref{fig:merge_gtmerge}), while the na\"{i}ve approach over-emphasises overlapping regions. Best viewed in colour.}
\label{fig:merge}
\end{figure*}

Algorithm~\ref{alg:GMMerge} outlines GMMerge, our efficient algorithm for parsimoniously approximating the merged mixture without weighting the intersection regions disproportionately. Each density of $\mathcal{G}_{\mathcal{X}}$ is re-weighted using a sparsity-inducing piecewise linear function. The parameter $t \in [0, \infty)$ controls how many densities are added. For $t = 0$, $\mathcal{G}_{\mathcal{XY}}$ contains only $\mathcal{G}_{\mathcal{Y}}$. As $t \to \infty$, $\mathcal{G}_{\mathcal{XY}}$ additionally contains every non-redundant density from $\mathcal{G}_{\mathcal{X}}$. Figure~\ref{fig:merge} shows the SVGM representations of two 2D point-sets, the na\"{i}vely merged mixture and the GMMerge mixture.

\begin{algorithm}
\begin{algorithmic}[1]
\Require aligned mixture models with unknown overlap $\mathcal{G}_{\mathcal{X}}$ and $\mathcal{G}_{\mathcal{Y}}$, parametrised by means $\boldsymbol{\mu}$, variances $\sigma^2$ and mixture weights $\phi$, and merging parameter $t$

\Ensure merged model $\mathcal{G}_{\mathcal{XY}}$

\State Initialise merged model: $\mathcal{G}_{\mathcal{XY}} \gets \mathcal{G}_{\mathcal{Y}}$

\For{$i = 1 \dots m$}

\State For the $i$th density of $\mathcal{G}_{\mathcal{X}}$, calculate:

\Statex[1] $\Delta = p \left( \boldsymbol{\mu}_{i,\mathcal{X}} \middle| \mathcal{G}_{i, \mathcal{X}} \right) - p \left( \boldsymbol{\mu}_{i,\mathcal{X}} \middle| \mathcal{G}_{\mathcal{Y}} \right)$

\State Update weight using sparsity-inducing function:

\Statex[1] $\phi_{i, \mathcal{X}} \gets \phi_{i, \mathcal{X}} \max \left( 0 , \min \left( 1, t \Delta \right) \right)$

\If{$\phi_{i, \mathcal{X}} > 0$}

\State Add to merged mixture: $\mathcal{G}_{\mathcal{XY}} \gets \mathcal{G}_{i, \mathcal{X}} \cdot \mathcal{G}_{\mathcal{XY}}$

\EndIf

\EndFor

\State Renormalise  $\mathcal{G}_{\mathcal{XY}}$

\end{algorithmic}
\caption{GMMerge: An algorithm for parsimonious Gaussian mixture merging}
\label{alg:GMMerge}
\end{algorithm}

\section{Experimental Results}
\label{sec:results}

SVR was tested using many different point-sets, including synthetic and real datasets in 2D and 3D, at a range of motion scales and outlier, noise and occlusion fractions.
In all experiments, the initial transformation parameter $\boldsymbol{\theta}$ was the identity, $\nu$ was 0.01 and $\gamma$ was selected by cross-validation, except where otherwise noted.
For all benchmark methods, parameters were chosen using a grid search.

\subsection{2D Registration}
\label{sec:results_2d}

\begin{figure}[!t]
\centering
\begin{subfigure}[]{0.49\columnwidth}
\includegraphics[height=2.0cm]{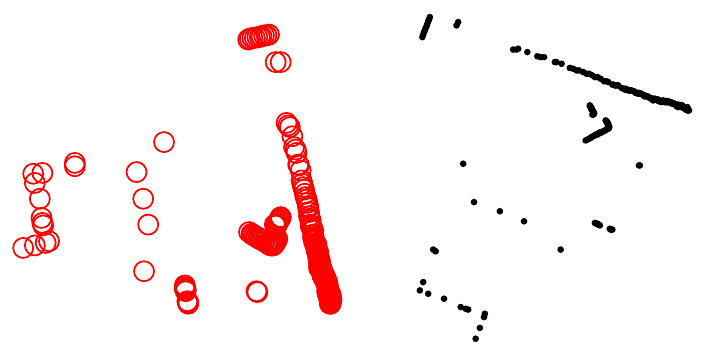}
\caption{\textsc{road} with rotation}
\label{fig:results_2D_road}
\end{subfigure}
\hfill
\begin{subfigure}[]{0.49\columnwidth}
\includegraphics[height=2.0cm]{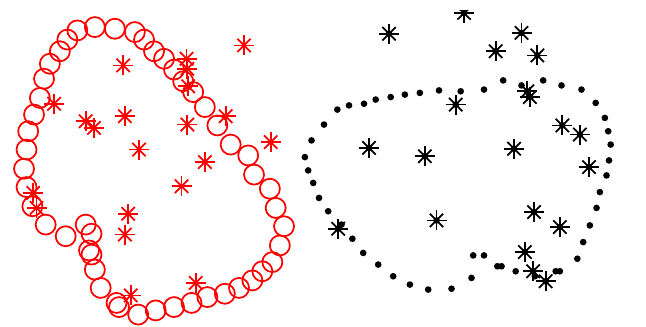}
\caption{\textsc{contour} with outliers}
\label{fig:results_2D_contour}
\end{subfigure}

\begin{subfigure}[]{0.49\columnwidth}
\includegraphics[height=2.0cm]{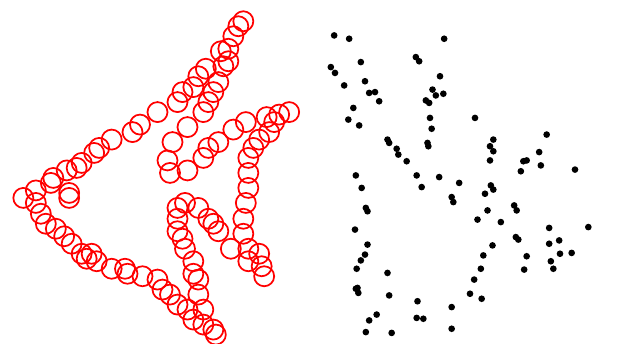}
\caption{\textsc{fish} with noise}
\label{fig:results_2D_fish}
\end{subfigure}
\hfill
\begin{subfigure}[]{0.49\columnwidth}
\includegraphics[height=2.0cm]{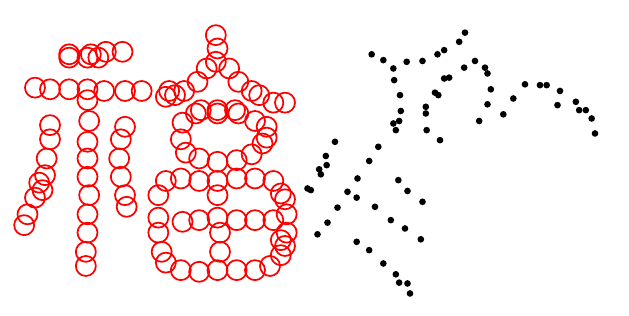}
\caption{\textsc{glyph} with occlusion}
\label{fig:results_2D_glyph}
\end{subfigure}
\caption{Sample scene (left) and model (right) point-sets from each 2D dataset, undergoing a range of perturbations.}
\label{fig:results_2D_datasets}
\end{figure}

To test the efficacy of SVR for 2D registration, the four point-sets in Figure~\ref{fig:results_2D_datasets} were used: \textsc{road}\footnote{Point-set from Tsin and Kanade \cite{tsin2004correlation}, available at \nolinkurl{http://www.cs.cmu.edu/~ytsin/KCReg/KCReg.zip}}, \textsc{contour}, \textsc{fish} and \textsc{glyph}\footnote{Point-sets from Chui and Rangarajan \cite{chui2003new}, available at \nolinkurl{http://cise.ufl.edu/~anand/students/chui/rpm/TPS-RPM.zip}}. Three benchmark algorithms were chosen: Gaussian Mixture Model Registration (abbreviated to GMR) \cite{jian2011robust}, Coherent Point Drift (CPD) \cite{myronenko2010point} and Iterative Closest Point (ICP) \cite{besl1992method}.
Annealing was applied for both SVR ($\delta = 10$) and GMR.
Note that the advantages of SVR manifest themselves more clearly on denser point-sets.

The range of motions for which a correct registration result was attained was tested by rotating the model point-set by $\alpha \in [-3.14, 3.14]$ radians with a step size of $0.01$.
In Table~\ref{tab:results_2d_motion_convergence_range}, we report the range of contiguous initial rotations for which the algorithm converged, chosen as a rotation error $\leq 1^{\circ}$. They show that SVR has a wider basin of convergence than the other methods, even for sparse point-sets.

\begin{table}[!t]
\centering
\caption{Convergence range (in radians). All rotation initialisations within these ranges converged (rotation error $\leq 1^{\circ}$).}
\label{tab:results_2d_motion_convergence_range}
\newcolumntype{C}{>{\centering\arraybackslash}X}
\begin{tabularx}{\columnwidth}{l C C C C}
\hline
\textbf{Point-Set} & \textbf{SVR} & \textbf{GMR} & \textbf{CPD} & \textbf{ICP}\\
\hline
\textsc{road} & \textbf{-3.1--3.1} & -3.0--3.0 & -1.6--1.6 & -0.8--0.8\\
\textsc{contour} & \textbf{-1.6--1.6} & -1.5--1.5 & -1.5--1.5 & -0.1--0.1\\
\textsc{fish} & \textbf{-1.6--1.6} & -1.5--1.5 & -1.2--1.3 & -0.4--0.5\\
\textsc{glyph} & \textbf{-1.6--1.6} & \textbf{-1.6--1.6} & -1.6--1.5 & -0.4--0.4\\
\hline
\end{tabularx}
\end{table}

\begin{figure}[!t]
\centering
\begin{subfigure}[]{0.49\columnwidth}
\includegraphics[width=\columnwidth]{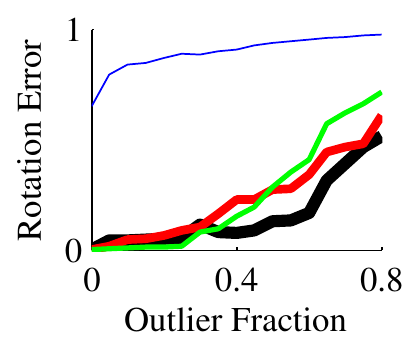}
\caption{Rotation error vs outlier fraction}
\label{fig:results_2D_outlier}
\end{subfigure}
\hfill
\begin{subfigure}[]{0.49\columnwidth}
\includegraphics[width=\columnwidth]{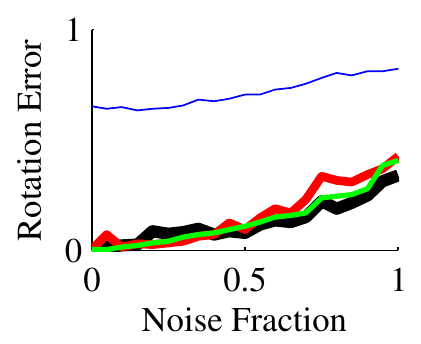}
\caption{Rotation error vs noise fraction}
\label{fig:results_2D_noise}
\end{subfigure}

\begin{subfigure}[]{0.51\columnwidth}
\includegraphics[width=\columnwidth]{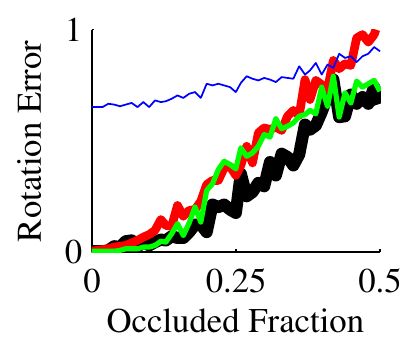}
\caption{Rotation error vs occluded fraction}
\label{fig:results_2D_occlusion}
\end{subfigure}
\hfill
\begin{subfigure}[]{0.48\columnwidth}
\includegraphics[width=\columnwidth, trim=0 0.2cm 0 0, clip]{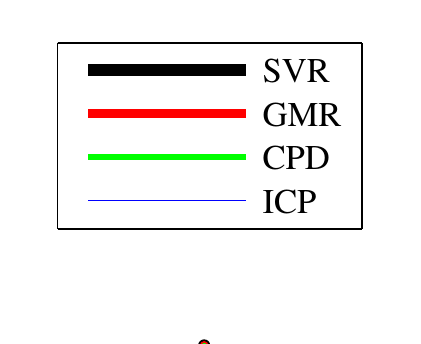}
\label{fig:results_2D_legend}
\end{subfigure}
\caption{Outlier, noise and occlusion results for the 2D point-sets. The mean rotation error (in radians) of 50 repetitions is reported for each and the results show that SVR is relatively robust to a large range of perturbations commonly found in real data.}
\label{fig:results_2D}
\end{figure}

To test the algorithm's robustness to outliers, additional points were randomly drawn from the uniform distribution and were concatenated with the model and scene point-sets separately. To avoid bias, the outliers were sampled from the minimum covering circle of the point-set. The motion was fixed to a rotation of $1$~radian ($57^{\circ}$) and the experiment was repeated $50$ times with different outliers each time. The mean rotation error for a range of outlier fractions is shown in Figure~\ref{fig:results_2D_outlier} and indicates that the proposed method is more robust to outliers than the others for large outlier fractions.

To test for robustness to noise, a noise model was applied to the model point-set by adding Gaussian noise to each point sampled from the distribution
$\mathcal{N} ( \mathbf{0}, ( \lambda \hat{\sigma} )^2 )$
, where $\lambda$ is the noise fraction and $\hat{\sigma}$ is the estimated generalised standard deviation across the entire point-set (\ref{eqn:sigma_hat}).
A fixed rotation of $1$~radian was used and the experiment was repeated $50$ times, resampling each time. The average rotation error for a range of noise fractions is shown in Figure~\ref{fig:results_2D_noise} and indicates that SVR is comparable to the other methods.

To test for robustness to occlusions, we selected a random seed point and removed a fraction of the model point-set using $k$-nearest neighbours.
A fixed rotation of $1$~radian was used and the experiment was repeated $50$ times with different seed points. The mean rotation error for a range of occlusion fractions is shown in Figure~\ref{fig:results_2D_occlusion} and indicates that the algorithm is more robust to occlusion than the others.

\subsection{3D Registration}
\label{sec:results_3d}

The advantages of SVR are particularly apparent with dense 3D point-sets. For evaluation, we used \textsc{dragon-stand}\footnote{Point-set from Brian Curless and Marc Levoy, Stanford University, at \nolinkurl{http://graphics.stanford.edu/data/3Dscanrep/}}, \textsc{aass-loop}\footnote{Point-set from Martin Magnusson, \"{O}rebro University, at \nolinkurl{http://kos.informatik.uni-osnabrueck.de/3Dscans/}} and \textsc{hannover2}\footnote{Point-set from Oliver Wulf, Leibniz University, at \nolinkurl{http://kos.informatik.uni-osnabrueck.de/3Dscans/}}
%
and seven benchmark algorithms: GMMReg (abbreviated to GMR) \cite{jian2011robust}, CPD \cite{myronenko2010point},  ICP~\cite{besl1992method}, NDT Point-to-Distribution (NDP) \cite{magnusson2007scan} and NDT Distribution-to-Distribution (NDD) \cite{stoyanov2012fast}, Globally-Optimal ICP (GOI) \cite{yang2013goicp} and \textsc{Super 4PCS} (S4P) \cite{mellado2014super}.
Annealing was used only where indicated.

To evaluate the performance of the algorithm with respect to motion scale, we replicated the experiment in \cite{jian2011robust} using the \textsc{dragon-stand} dataset. This contains 15 self-occluding scans of the dragon model acquired from different directions.
We registered all 30 point-set pairs with a relative rotation of $\pm 24^{\circ}$ and repeated this for $\pm 48^{\circ}$, $\pm 72^{\circ}$ and $\pm 96^{\circ}$. As per \cite{jian2011robust}, the criterion for convergence was $\hat{q} \cdot q > 0.99$, where $\hat{q}$ and $q$ are the estimated and ground truth quaternions respectively.
While $\gamma$ was selected by cross-validation, using the estimate $\hat{\sigma}$ yielded a very similar result.
The number of correctly converged registrations is reported in Table~\ref{tab:results_3d_motion_fraction}, showing that SVR has a significantly larger basin of convergence than the other local methods and is competitive with the slower global methods.

\begin{table}[!t]
\centering
\caption{Number of point-set pairs that converged for a range of relative poses. Mean computation time in seconds is also reported.}
\label{tab:results_3d_motion_fraction}
\newcolumntype{C}{>{\centering\arraybackslash}X}
\begin{tabularx}{\columnwidth}{C C C C C | C C}
\hline
& \multicolumn{4}{c|}{\textbf{Local}} &  \multicolumn{2}{c}{\textbf{Global}}\\
\textbf{Pose} & \textbf{SVR} & \textbf{GMR} & \textbf{CPD} & \textbf{ICP} & \textbf{GOI} & \textbf{S4P}\\
\hline
$\pm 24^{\circ}$ & \textbf{30} & 29 & 26 & 28 & \textbf{30} & 29\\
$\pm 48^{\circ}$ & \textbf{29} & 20 & 18 & 19 & 27 & 24\\
$\pm 72^{\circ}$ & 16 & 13 & 14 & 13 & \textbf{18} & 17\\
$\pm 96^{\circ}$ & 4 & 2 & 3 & 1 & 10 & \textbf{13}\\
\hline
Runtime & 0.2 & 19.2 & 5.7 & \textbf{0.04} & 1407 & 399\\
\hline
\end{tabularx}
\end{table}

\begin{figure}[!t]
\centering
\includegraphics[width=\columnwidth]{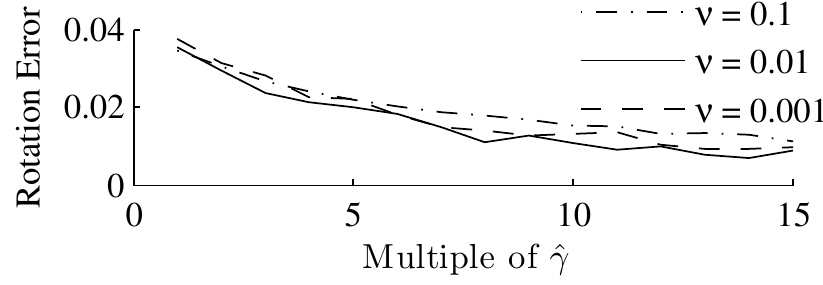}
\caption{Sensitivity analysis for $\gamma$ and $\nu$. The median rotation error (in radians) of all \textsc{dragon-stand} point-sets with $\pm 24^{\circ}$ pose differences are plotted with respect to multiples of $\hat{\gamma} = \sfrac{1}{2\hat{\sigma}^2}$.}
\label{fig:results_3D_sensitivity}
\end{figure}

A representative sensitivity analysis is shown in Figure~\ref{fig:results_3D_sensitivity} for the \textsc{dragon-stand} dataset. It indicates that rotation error is quite insensitive to perturbations in $\gamma$ and is very insensitive to $\nu$, justifying the choice of fixing this parameter.

\begin{figure}[!t]
\centering
\begin{subfigure}[]{0.49\columnwidth}
\includegraphics[width=\columnwidth]{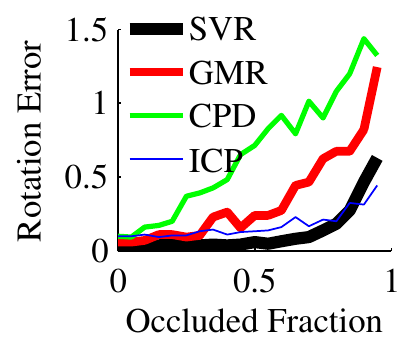}
\caption{Mean rotation error for $\pm 24^{\circ}$}
\label{fig:results_3D_occlusion_mean_24}
\end{subfigure}
\hfill
\begin{subfigure}[]{0.49\columnwidth}
\includegraphics[width=\columnwidth]{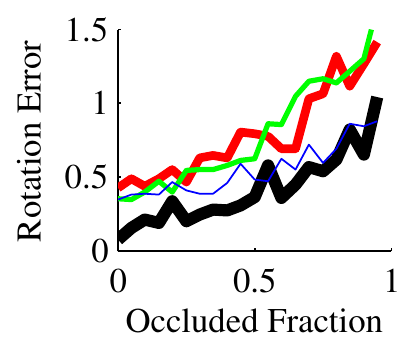}
\caption{Mean rotation error for $\pm 48^{\circ}$}
\label{fig:results_3D_occlusion_mean_48}
\end{subfigure}

\begin{subfigure}[]{0.49\columnwidth}
\includegraphics[width=\columnwidth]{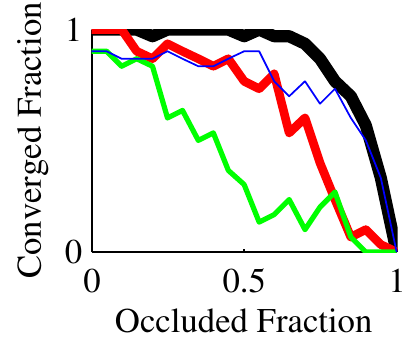}
\caption{Convergence rate for $\pm 24^{\circ}$}
\label{fig:results_3D_occlusion_nSuccesses_24}
\end{subfigure}
\hfill
\begin{subfigure}[]{0.49\columnwidth}
\includegraphics[width=\columnwidth]{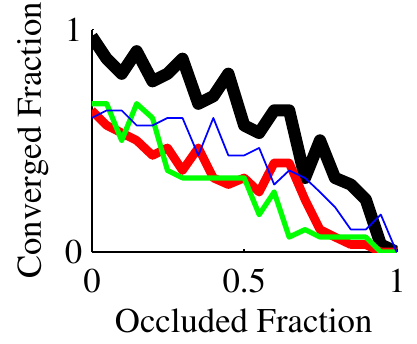}
\caption{Convergence rate for $\pm 48^{\circ}$}
\label{fig:results_3D_occlusion_nSuccesses_24}
\end{subfigure}
\caption{Mean rotation error (in radians) and convergence rate of all \textsc{dragon-stand} point-sets with $\pm 24^{\circ}$ and $\pm 48^{\circ}$ pose differences, with respect to the fraction of occluded points.}
\label{fig:results_3D_occlusion}
\end{figure}

To evaluate occlusion robustness, the same procedure was followed as for 2D, using the \textsc{dragon-stand} dataset. The mean rotation error (in radians) and the fraction of correctly converged point-set pairs with respect to the fraction of occluded points is shown in Figure~\ref{fig:results_3D_occlusion}, for relative poses of $\pm 24^{\circ}$ and $\pm 48^{\circ}$. The results show that SVR is significantly more robust to occlusion than the other methods.

\begin{figure}[!t]
\centering
\begin{subfigure}[]{0.66\columnwidth}
\includegraphics[width=\columnwidth]{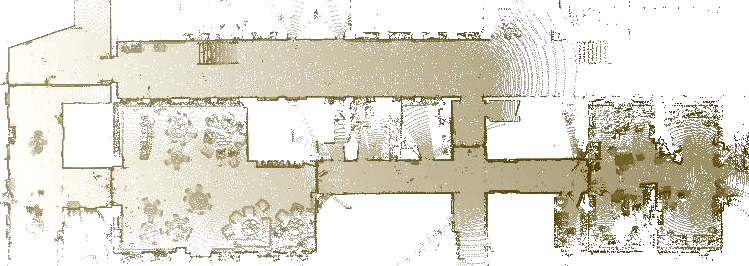}
\caption{\textsc{aass-loop}}
\label{fig:results_3D_aerial_aass}
\end{subfigure}
\hfill
\begin{subfigure}[]{0.32\columnwidth}
\includegraphics[width=\columnwidth]{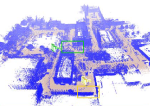}
\caption{\textsc{hannover2}}
\label{fig:results_3D_aerial_hannover2}
\end{subfigure}
\caption{Two large-scale 3D datasets.}
\label{fig:results_3D_aerial}
\end{figure}

Finally, we report registration results on two large real-world 3D datasets shown in Figure~\ref{fig:results_3D_aerial}: \textsc{aass-loop} ($60$ indoor point-sets with ${\sim}13\,500$ points on average) and \textsc{hannover2} ($923$ outdoor point-sets with ${\sim}10\,000$ points on average), after downsampling using a 0.1~m grid. Both were captured using a laser scanner and ground truth was provided.
%
%
These are challenging datasets because sequential point-sets overlap incompletely and occluded regions are present.
The results for registering adjacent point-sets are shown in Table~\ref{tab:results_3d_large_aass} for \textsc{aass-loop} and Table~\ref{tab:results_3d_large_hannover2} for \textsc{hannover2}. The ICP and annealed NDT results are reported directly from Stoyanov~\etal \cite{stoyanov2012fast} and we use their criteria for a successful registration (inlier): a translation error less than 0.5~m and a rotation error less than 0.2~radians. SVR outperforms the other methods by a significant margin, even more so when annealing ($\delta = 2$) is applied (SVR\textsuperscript{+}).

\begin{table}[!t]
\centering
\caption{Registration results for \textsc{aass-loop}. While mean translation error (in metres) and rotation error (in radians) are commonly reported, the percentage of inliers (successful registrations) is a more useful metric for comparison. The mean computation time (in seconds) is also reported. SVR\textsuperscript{+} is SVR with annealing.}
\label{tab:results_3d_large_aass}
\newcolumntype{C}{>{\centering\arraybackslash}X}
\begin{tabularx}{\columnwidth}{l C C C C C C C}
\hline
\textbf{Metric} & \textbf{SVR} & \textbf{SVR\textsuperscript{+}} & \textbf{GMR} & \textbf{ICP} & \textbf{NDP} & \textbf{NDD} & \textbf{S4P}\\
\hline
Transl. & 0.95 & \textbf{0.67} & 1.61 & 0.99 & 1.10 & 0.85 & 0.71\\
Rotation & 0.08 & 0.06 & 0.12 & 0.04 & \textbf{0.02} & 0.06 & 0.32\\
Inlier \% & 81.4 & \textbf{86.4} & 18.6 & 55.2 & 50.0 & 63.8 & 78.0\\
\hline
Runtime & 3.43 & 29.7 & 599 & 10.8 & 9.12 & \textbf{1.02} & 60.7\\ 
\hline
\end{tabularx}
\end{table}

\begin{table}[!t]
\centering
\caption{Registration results for \textsc{hannover2}. The mean translation error (in metres), rotation error (in radians), inlier percentage and mean runtime (in seconds) are reported. SVR\textsuperscript{+} uses annealing.}
\label{tab:results_3d_large_hannover2}
\newcolumntype{C}{>{\centering\arraybackslash}X}
\begin{tabularx}{\columnwidth}{l C C C C C C C}
\hline
\textbf{Metric} & \textbf{SVR} & \textbf{SVR\textsuperscript{+}} & \textbf{GMR} & \textbf{ICP} & \textbf{NDP} & \textbf{NDD} & \textbf{S4P}\\
\hline
Transl. & 0.10 & \textbf{0.09} & 1.32 & 0.43 & 0.79 & 0.40 & 0.40\\
Rotation & \textbf{0.01} & \textbf{0.01} & 0.05 & 0.05 & 0.05 & 0.05 & 0.03\\
Inlier \% & \textbf{99.8} & \textbf{99.8} & 8.88 & 74.4 & 54.2 & 76.4 & 75.0\\
\hline
Runtime & 14.0 & 32.6 & 179 & 5.68 & 4.03 & \textbf{0.51} & 39.7\\
\hline
\end{tabularx}
\end{table}

The mean computation speeds of the experiments, regardless of convergence, are reported in Tables~\ref{tab:results_3d_motion_fraction}, \ref{tab:results_3d_large_aass} and~\ref{tab:results_3d_large_hannover2}. All experiments were run on a PC with a 3.4~GHz Quad Core CPU and 8~GB of RAM.
%
%
The SVR code is written in unoptimised MATLAB, except for a cost function in C++, and uses the LIBSVM \cite{chang2011libsvm} library. The benchmarking code was provided by the respective authors, except for ICP, for which a standard MATLAB implementation with k-d tree nearest-neighbour queries was used.
For the \textsc{dragon-stand} speed comparison, all point-sets were randomly downsampled to $2\,000$ points, because GMR, CPD, GOI and S4P were prohibitively slow for larger point-sets.

\section{Discussion}
\label{sec:discussion}

The results show that SVR has a larger region of convergence than the other methods and is more robust to occlusions. This is an expected consequence of the SVGM representation, since it is demonstrably robust to missing data.
In addition, the computation time results show that it scales well with point-set size, unlike GMR and CPD, largely due to the data compression property of the one-class SVM. There is a trade-off, controlled by the parameter $\gamma$, between registration accuracy and computation time.

For the application of accurate reconstruction using our framework, the one-class SVM may be replaced with a two-class SVM to better model the fine details of a scene.
To generate negative class (free space) training points, surface points were displaced along their approximated normal vectors by a fixed distance $d$ and then those points that were closer than $0.9d$ to their nearest surface point were discarded. The SVGMs constructed using this approach may be fused using GMMerge. However, for the purposes of registration, capturing fine detail in this way is unnecessary, counter-productive and much less efficient.

While SVR is a local algorithm, it can still outperform global algorithms on a number of measures, particularly speed, for certain tasks.
In Section~\ref{sec:results_3d}, we compared SVR with the guaranteed-optimal method Globally-Optimal ICP (GOI)~\cite{yang2013goicp} and the faster but not optimal method \textsc{Super 4PCS} (S4P) \cite{mellado2014super}. The motion scale results of GOI were comparable to our method, while the average runtime was four orders of magnitude longer. Note that, for point-sets with missing data or partial overlap, a globally-optimal alignment is not necessarily correct. S4P had a more favourable runtime--accuracy trade-off but was nonetheless outperformed by SVR.

\section{Conclusion}
\label{sec:conclusion}

In this paper, we have presented a framework for robust point-set registration and
%
%
merging using a continuous data representation. Our point-set representation is constructed by training a one-class SVM and then approximating the output function with a GMM.
This representation is sparse
and robust to occlusions and missing data, which are crucial attributes for efficient and robust registration.

The central algorithm, SVR, outperforms state-of-the-art approaches in 2D and 3D rigid registration, exhibiting a larger basin of convergence. In particular, we have shown that it is robust to occlusion and missing data and is computationally efficient.
The GMMerge algorithm complements the registration algorithm by providing a parsimonious and equitable method of merging aligned mixtures, which can subsequently be used as an input to SVR.

There are several areas that warrant further investigation. Firstly, there is significant scope for optimising the algorithm using, for example, approximations like the improved fast Gauss Transform \cite{yang2003improved} or faster optimisation algorithms that require an analytic Hessian. Secondly, non-rigid registration is a natural extension to this work and should benefit from the robustness of SVR to missing data.
It may also be useful to train the SVM with full data-driven covariance matrices \cite{abe2005training} and use the full covariances for registration \cite{stoyanov2012fast}.
Finally, methods of constructing tight bounds for an efficient branch-and-bound framework based on SVR could be investigated in order to implement a globally-optimal registration algorithm.

{\small
\bibliographystyle{ieee}
\bibliography{citations_abbrv}
}

\end{document}